%% file: main.tex
\documentclass[letterpaper]{article}
\usepackage{aaai}
\usepackage{times}
\usepackage{helvet}
\usepackage{courier}
\usepackage{booktabs}
\usepackage{textcomp}
\usepackage{url}
\usepackage{subcaption}
\usepackage{tabularx}
\usepackage{graphicx}
\usepackage{xcolor}
\usepackage[margin=1in]{geometry}
\usepackage{graphicx, blindtext}
\usepackage{algorithm2e}

\usepackage{amsmath}

\usepackage{float}
\restylefloat{table}
\usepackage{placeins}

\usepackage{mathtools, nccmath}
\usepackage{xparse}
\DeclarePairedDelimiterX{\set}[1]{\{}{\}}{\setargs{#1}}
\NewDocumentCommand{\setargs}{>{\SplitArgument{1}{;}}m}
{\setargsaux#1}
\NewDocumentCommand{\setargsaux}{mm}
{\IfNoValueTF{#2}{#1} {#1\,\delimsize|\,\mathopen{}#2}}

\title {Efficient Document Image Classification Using Region-Based Graph Neural Network}

\author {Jaya Krishna Mandivarapu, Eric Bunch, Qian You, Glenn Fung \\
American Family Insurance, Machine Learning Research Group \\
jmandivarapu1@student.gsu.edu , \{ebunch, qyou, gfung\}@amfam.com \\
}
\begin{document}
%
\maketitle
\input{main/abstract}
\input{main/introduction}
\input{main/rel_work}

\input{main/methods_copy}

\input{main/datasets_and_hyperparameters}
\input{main/experiments_and_results}
\input{main/deployment}

\input{main/conclusion_and_nextsteps}
\bibliographystyle{aaai}
\bibliography{main.bib}
\end{document}

%% file: main/abstract.tex
\begin{abstract}
Document image classification remains a popular research area because it can be commercialized in many enterprise applications across different industries. Recent advancements in large pre-trained computer vision and language models and graph neural networks has lent document image classification many tools. However using large pre-trained models usually requires substantial computing resources which could defeat the  cost-saving advantages of  automatic document image classification. In the paper we propose an efficient document image classification framework that uses graph convolution neural networks and incorporates textual, visual and layout information of the document.
 Empirical results on both publicly available and real-world data show that our methods achieve near SOTA performance yet require much less computing resources and time for model training and inference. This results in solutions than offer better cost advantages, especially in scalable deployment for enterprise applications.  
\end{abstract}

%% file: main/introduction.tex
\section{Introduction}
\begin{figure*}[t]
  \begin{center}
    \includegraphics[width=0.9\textwidth]{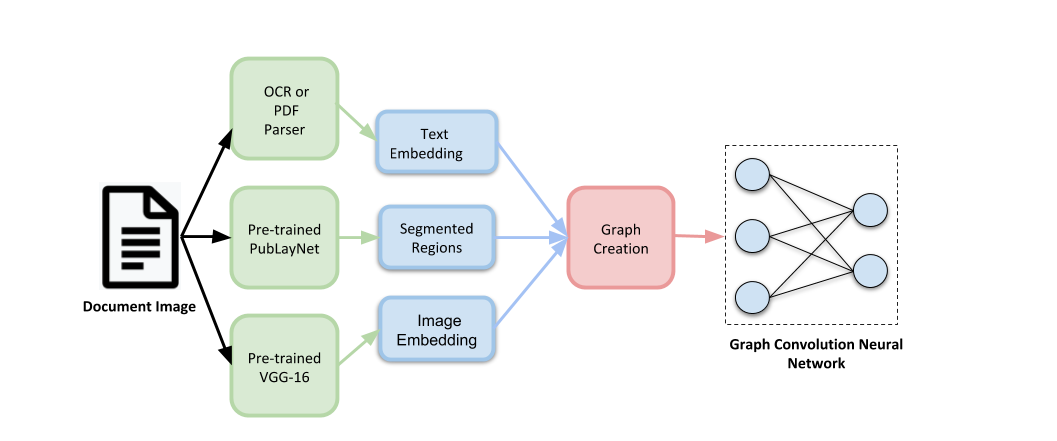}
  \end{center}
  \caption{\textbf{Eff-GNN Framework overview:} textual embedding, segmented regions and image embeddings of an image are integrated when the graph of document is formed. Created graph is fed into the Graph Convolution Neural Network for graph classification as document image classification.}
  \label{fig:overview}
\end{figure*}

Gartner has estimated 80\% of enterprises data is unstructured (emails, PDF and other documents). These documents contain rich information and knowledge about internal and external business communication and transactions. And they have ubiquitous applications in numerous industrial sectors such as finance, health care, and law etc. Therefore, being able to automatically and efficiently sort, analyze, and extract structure and content from document images can improve  efficiency and reduce cost for many business workflows. Document image classification is an import task in these automation solutions, and has been a popular research area for decades. Early works usually build classifiers that rely on Optical Character Recognition (OCR) to extract text information, and employ heuristics to model layout structural features. In light of the advancement of computer vision and deep learning, VGG-16 \cite{VGG16} pre-trained on ImageNet \cite{imagenet_cvpr09} reported good classification performance on data sets mixed of business letters, print advertisement, emails and magazine articles \cite{kumar2014structural}. Both \cite{bertgrid} and \cite{layoutLM} created document representations by encoding layout coordinates into positional embeddings as inputs to pre-trained BERT \cite{BERT} or transformer architectures. The latest PubLayNet \cite{publaynet} addresses the limited public available document image data sets by training a Mask R-CNN \cite{maskrcnn} model on 360k images of scientific articles, and enables transfer learning to other document domains. Motivated by the development of graph neural network algorithms \cite{gnn_survey,dgcnn,gnn_diffpool}, researchers \cite{gnn_multimodal} attempted to use graph convolutions to model the interactions among structural components of a document and between the visual and textual features, as an alternative to pixel level or token level document modeling. 
In contrast to fast moving research progress in document analysis and classification, few have systematically studied the time and hardware resources when using different methods and the financial implications of the model design. However, as document image classifications have been primarily motivated by its potential in commercialization, it is imperative to study its model performance with computing resources requirements and financial implications. In this paper we propose an efficient document image classification framework as shown in Fig. \ref{fig:overview}. Semantics regions of a document is extracted by pre-trained PubLayNet, textual features are extracted by text embedding models and the image features are extracted by a pre-trained VGG-16 model. Graphs formed for the document, with the document class labels are used to train a sort pooling graph convolution network \cite{dgcnn} which normalizes and classify arbitrary graphs therefore documents. 
The major contributions of our papers are as follows:
\begin{itemize}
    \item We propose a novel document image classification framework which applies a graph convolution neural network to a document image graph formed by semantic regions extracted from a pre-trained document segmentation model. Moreover both image and text features of the regions are extracted and assigned to the nodes so that information from both modalities are captured and propagated in the graph convolutions. To our best knowledge, our framework is the the first in effectively and economically integrating image, text, and layout information for document image classification using a graph convolution neural network.
    \item We have rigorously bench marked our proposed method against state-of-the-art pre-trained vision models and transformer language models on document image data sets. These include an insurance related document image data set consisted of 11 classes and an open source data set of 10 classes. The results showed the classification results of our method are comparable to those of baseline models, if not better. 
    \item We  extensively bench marked the computing resources required by all methods. The results showed our framework needs substantially less computing resources and less time, further indicating the cost advantages of training, deployment and hosting at scale. Efficient model also helps accelerate model iterations and update.   
\end{itemize}
We also discussed a few potential document image classification applications and the infrastructure to deploy our framework. The potentially large scale adoption of document image classification further reinforced the need for an efficient document image classification method.  

%% file: main/rel_work.tex
\section{Related Work}
Early document image classification algorithms relied on OCR to extract content information and exploited the visual structure and layout of a document image, e.g. using tree-related data structures to model a document \cite{Dengel93,ShinDR01,DiligentiFG03}. The subsequent decades of research document analysis, including document image classification, evolves around more sophisticated ways to leverage the image features, text features and document layout information. \par  
Advancement in Deep Convolutional Neural Networks (DCNN) lend new tools for document 
image classification \cite{CNNDOC}, because DCNN could extract salient and hierarchical visual feature representations which can somewhat reflect hierarchical nature of document layout. Quite a few DCNN training strategies for document image classification are proposed and extensively reviewed \cite{CNNDOC,cutting_error_half}. Variants of VGG-16 \cite{DCGNN} achieved the state-of-the art on publicly available Tobacco data sets \cite{LewisAAFGH06}.
Document understanding and analysis community have also been leveraging word embedding techniques \cite{word2vec}  in NLP and large language models \cite{BERT} to create contextualized embedding for textual content in an document image. BERTGrid \cite{bertgrid} uses both the contextualized word embeddings and its 2D layout coordinates to extract information by predicting segmentation masks and bounding boxes. Assuming syntactic features matter less than content categories in document classification. DocBERT achieved an economical solution for document classification task by distilling BERT \cite{DevlinCLT19}.  LayoutLM \cite{layoutLM} jointly models interactions between text and layout information by inputting both text embeddings along with its 2D layout positional embeddings extracted using OCR and Region of Interest (ROI) Regressions. Considering both the image and textual modalities in the document images, multi-modalities methods \cite {multimodal_classification,end2end_multimodal_extraction} are adapted to document classification tasks as well. \par

%% file: main/methods_copy.tex
\section{Methodology}

In this section, we briefly review the advantages and limitations when using either CNNs or large language models in document classification. We also discuss document segmentation and the intuitions of using region based representations. We then describe our proposed efficient graph neural network, \textbf{Eff-GNN}  
\subsection{Deep Convolution Neural Network Learning Approaches}
When using deep convolutional neural networks for document image classification, the document is treated as an image and is ingested as tensor representing the pixel values of the image. VGG-16 pre-trained on ImageNet \cite{DengDSLL009} can achieve good results on general business documents \cite{DCGNN}. Even for the insurance data set, VGG-16 pre-trained on ImageNet can be a powerful visual feature extractor.


\subsection{Language Model based Approaches}
In general, BERT-like pre-trained language models achieve superior performance on natural language processing tasks by adding self attention mechanism and positional information to the encoder-decoder architecture. When using BERT-like models to classify document images, we classify based on the contextualized embedding of text extracted from image. DocBert \cite{DOCBERT} assumes syntax features matter less if only the categories of the document need to be decided. And DocBert successfully distills trained BERT into a much smaller LSTM model. This gives us some insight that token level modeling for document classification may not be necessary.   

\subsection{Document Segmentation}
A business document contains visually salient structural components such as header, footer, paragraph, table etc. Intuitively one can classify a document image by its layout and structural components without accessing much of its content. The regions of structural components are usually pixels or tokens with similar appearances or groupings. Hence regions are a higher level abstraction and representation which we can leverage to classify this document. Research in the computer vision community has provided plenty of tools of segmenting document images. The latest advancement is PubLayNet which trained a Mask R-CNN model for 360 thousand document images from scientific articles. The segmented region results from PubLayNet can be found at Figure 
\ref{fig:insurance_imgs} (c) \& (d).

 
\subsection{Efficient GNN for Document Image Classification}
Ideally an effective document classification method need to leverage both textual, image and layout information. However, training or fine tuning CNNs or large language models do not only run into resource constraints (e.g. GPUs, memory ), but also prevent fast model iterations. We attempted to address this dilemma by graph representations using graph representations to represent document. We then assign image and text features to the nodes of the graph and apply a graph convolution neural network. Finally we classify the document as classifying a graph. \par


Details of training our proposed \textbf{Eff-GNN} can be found in Algorithm 1. For each image with class label, we extract its text using OCR PyTesseract and we extract its semantic regions using PubLayNet. Each image is converted into a graph where each node is a region. To generate the text feature of the nodes, we use Word2Vec to create the embeddings for words in the region; to generate image feature of the nodes, we extract visual features from that region using VGG-16 pre-trained on ImageNet. Text features and image features of the node can be concatenated and assigned to the nodes. A graph convolution neural network classifier with a SortPooling \cite{dgcnn} layer is then trained on this data. We adapted to this specific graph convolution neural network because it preserve features of individual nodes and also enforces learning from graph global topology. \par 
Till this end we have integrated textual, image and layout information into a document classification task using a graph convolution neural network.

\RestyleAlgo{algoruled}
 \begin{algorithm}[h!]
 \caption{Efficent Graph Neural Network Training}
        \DontPrintSemicolon
        \SetKwInput{kwInput}{Input}
        \SetKwBlock{kwInit}{Preprocessing}{end}
        \SetKwBlock{kwMain}{Training}{end}
        \SetKwInput{kwOutput}{Output}
        \kwInput{
            $N$ Training documents $\{(D_1 , y_1 ), \dots , (D_N , y_N )\} $.}
        \kwInit{
        \par
        \vspace{0.25cm}
        
            \For{$m = 1, \dots , N$}{
                Extract textual information $t_m$, detect layout regions $s_m$ and visual features $I_m$ present in $D_m$\;
            }
            
    Train word2vec  $\mathcal{W}$=$word2vec$(C)
    \[ C = \set{t_i ; 1\leq i \leq N} \]%
       \For{$m = 1,\dots , N$}{
               Convert each document ($D_m$) into graph ($G_m$) using  $t_m$,$s_m$,  $I_m$ and $\mathcal{W}$ which would be $D_m$$\rightarrow$$G_m$  
            }
        
        \vspace{0.25cm}
        }
        \kwMain{
        \par
        \vspace{0.25cm}
            Training GNN on graph data \( \mathcal{Z}=\left\{\left(G_{1}, y_{1}\right),\left(G_{2}, y_{2}\right), \ldots\right\} \) where \( y_{i} \in \mathcal{Y} \) is the label corresponding to graph \( G_{i} \in \mathcal{G}, \) with goal of learning a mapping \( f: \mathcal{G} \rightarrow \mathcal{Y} \) that maps graphs to the set of labels
        \vspace{0.25cm}
        }
        \label{psuedocode}
    \end{algorithm}

%% file: main/datasets_and_hyperparameters.tex
\section{Experimental Setup}
\subsection{Datasets} 
\begin{figure*}[t]
\begin{center}
\def\tabularxcolumn#1{m{#1}}
\begin{tabularx}{\linewidth}{@{}cXX@{}}
\begin{tabular}{cccc}
\subfloat[Medical Record]{\includegraphics[width=3.5cm]{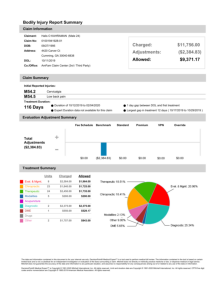}} 
   & \subfloat[Subrogation Letter]{\includegraphics[width=3.5cm]{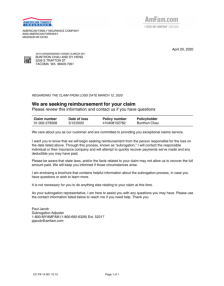}}
   & \subfloat[Medical Authorization]{\includegraphics[width=3.5cm]{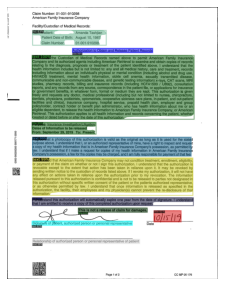}}
   & \subfloat[Attorney Correspondence]{\includegraphics[width=3.5cm]{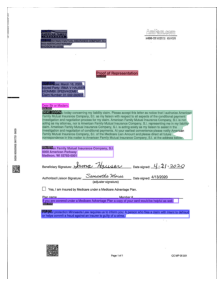}} \\
\end{tabular}
\end{tabularx}
\end{center}
\caption{Examples of documents from the Insurance data set. The document classes (a) Medical Record, (b) Subrogation Letter, (c) Medical Authorization, and (d) Attorney Correspondence. The two images (c) and (d) include a visualization of the output of the PubLayNet model.}
\label{fig:insurance_imgs}
\end{figure*}

We use the following two datasets to evaluate our proposed model: Insurance dataset (Fig.\ref{fig:insurance_imgs}), Tobacco-3482 dataset\cite{kumar2014structural}. Insurance dataset contains 5772 document images which spans across 11 categories and Tobacco-3482 dataset consists of 3482 images which spans across 10 categories. Categories from the Tobacco dataset include Advertisements, Emails, Memos, and Scientific Reports. Categories from the Insurance dataset include Medical Bills, Medical Authorizations, Medial Records, Police Reports, and Subrogation Letters. See Fig. \ref{fig:insurance_imgs} for examples of documents from the Insurance dataset.

For the Insurance dataset, we use the  
splits of 4544 images for train set and 1280 images  for test set; for Tobacco-3482 dataset, we use the
standard splits of  2482 images for training, 800 images for testing, and rest 200 images for validation set.


We also summarized the statistics of graphs that those documents formed in Table \ref{Table1:DatasetState}. As evidenced in the tables, the graph size of each document is significantly smaller relative to the number of pixels or the number of words in a document, which significantly simplifies the subsequent modeling and computation. 
\begin{table}
\begin{tabular}{p{1.1cm}p{0.15cm}p{0.15cm}p{0.15cm}p{0.15cm}p{0.15cm}p{0.15cm}p{0.15cm}p{0.15cm}p{0.15cm}p{0.15cm}p{0.15cm}}
\toprule
 class label & 0  &  1  &  2  &  3  &  4  &  5  &  6  &  7  &  8  &  9  &  10 \\
\midrule
  Insurance & 13 &   5 &   6 &  18 &   5 &  12 &   3 &  11 &  8 &  13 &  11 \\
\midrule
 Tobacco & 5 &  9 &  3 &  11 &  10 &  5 &  2 &  9 &  10 &  2 & - \\
\bottomrule
\end{tabular}
\caption{\textbf{Median number of nodes per class for the Insurance and Tobacco data sets.}}
\label{Table1:DatasetState}
\end{table}

\subsection{Document Pre-processing}
To construct a graph for each document image we need to utilize the different layout regions' information such as paragraphs, title, list, table present in each document. We process the scanned document images using pre-trained PubLayNet to obtain the necessary bounding boxes for each region. To extract the textual information present in the bounding box of each region in the document images we apply PyTesseract, an open-source python OCR package. All the results of the PubLayNet and PyTesseract were then serialized and stored in tensor format using PyTorch. \cite{Pytorch}. 

\subsection{Hyper parameters and infrastructure}
We use Hedwig\footnote{https://github.com/castorini/hedwig},
an open-source deep learning
toolkit with a number of implemented of document classification models. 
We use a Tesla K80 GPU for all models requiring GPU for train, and use amazon EC2-t2.micro and EC2-c type machine when only CPU is needed . We use PyTorch 1.5 as the backend framework, and gensim \cite{gensim} package for computing the node feature vectors using word2vec.\\





%% file: main/experiments_and_results.tex
\section{Results and Discussions}


\subsection{Comparing classification accuracy} 
We compare our proposed approach with the state-of-the-art deep learning models, VGG-16 pre-trained on ImageNet and $\text{BERT} _{base}$ pre-trained on Wikipedia. When using $\text{BERT} _{base}$ for classification, we extracted tokens from document images as input and fine tune $\text{BERT} _{base}$ with class labels. When using pre-trained VGG-16, we used it directly to extract document image features for classification. No further fine-tuning is used. We also compared our model with DocBERT which is specially designed for document level classification by simplifying BERT models with knowledge distillation. \par
Table \ref{effgnn_results_accuracy} shows the model comparison results of classification AUC on the two data sets. In the insurance data set, our proposed model achieves 90.7\% to 91.0 \%, very competitive as compared to models in BERT families (91.95 \%) and VGG-16 (90.6 \%). 
In the Tobacco-3482 data set, our models achieved 73.5 \% to 77.5 \% , comparable to models in BERT families (82.3 \%) and VGG-16 (81.5 \%). The fact our proposed model shows more advantages when classifying the insurance data set could be due to the high intra-class variance and low inter-class variance in the Tobacco-3482 data set \cite{KolschAEL17}. \par
We also experimented with combining text and image embedding features as node features in the graph neural network.  The combination does provides ample AUC improvements in our proposed method on Tobacco-3482 dataset  i.e. 73.5 \% for Eff-GNN + Word2Vec and for 77.5 \% for Eff-GNN + Word2Vec + Image Embedding. In the insurance data set, Eff-GNN +  Word2Vec + Image Embedding shows little improvement over using Word2Vec text features alone. That could be due to the fact that both the textual and image content in the insurance data set provides enough information for classification. This assumption can be further justified by the similar classification results achieved by models in BERT family and VGG-16.\par

In addition to classification performance, we compare the number of trainable parameters of each model. In both the insurance data set and Tobacco-3482, our model size is drastically smaller than models in BERT families and VGG-16. We calculate the parameters of our model as the sum of parameters in the graph neural network and the parameters in trained word2vec. The small sizes of graph neural network model (Table \ref{Table1:DatasetState}) results in only 160,000 parameters. The Word2Vec model is also relatively light weight because each of the data set contains a very limited vocabulary. Note we did not include the 44.2 million parameters of PubLayNet, because in our framework we do not train or fine-tune any parameters of the PubLayNet.


\subsection{Comparing computing resources}
We also bench marked the time and memory required for training our proposed Eff-GNN against other models. Table \ref{effgnn_results_resources_insurance} reports the statistics on the insurance data set, 4544 images for training and 1280 images for inference. Eff-GNN models take less than 5 minutes to train 50 epochs whereas VGG-16 or models in BERT Family take hours to train less number of epochs (15 epochs and 28 epochs respectively). In particular Eff-GNN can run on CPU alone and its model training time is comparable to its GPU counterpart. This is consistent with the small size of our model (See Table \ref{effgnn_results_accuracy},  "Parameters" column). Eff-GNN can achieve these advantages because it models the documents using a graph formed by regions extracted by PubLayNet. The time of using PubLayNet to extract regions for training images are negligible. The size of the resulting graph leads to a small model compared to deep models trained on pixel level information or transformers trained on token level information. Therefore Eff-GNN only uses 470MB in GPU memory with additional 3.5 GB for using PubLayNet. Consequently Eff-GNN requires drastically less time for the inference of 1280 images (0.79 seconds) as compared VGG-16 (103 seconds) and BERT (40 seconds). Note the time of document pre-porcessing steps such as OCR and training Word2Vec model are not included in the table. Although these two steps are extra for our proposed framework, we contend that their addition does not nullify the efficiencies gained through graph neural nets. Even BERT based models require OCR extraction pre-processing step. Just one Word2Vec needs to be trained for the entire data sets and OCR can be optimized by e.g. parallel processing.\par
Compared with the SOTA pre-trained large models, our proposed Eff-GNN framework achieved competitive classification results on our insurance document image data sets, and achieved comparable results on the the open source Tobacco-3482 data set. We also showed that combining text and image information as the node features in our graph neural network can be advantageous when OCR fails to extract text information or when the two modalities are complimentary. Our proposed method models document representations using extracted semantic regions, instead of using token level or pixel level information. Therefore our model size is dramatically less than other methods, and can be run on CPU machines.
\begin{table*}[ht]
\centering
\begin{tabular}{clcc}
\toprule
     \textbf{Data Set} &  \textbf{\hspace{70pt} Model} &  \textbf{AUC} & \textbf{\# Parameters}\\
\midrule
 \textbf{Insurance} & DocBert \cite{DOCBERT} & 91.95 \% & 110M \\
             & BERT \cite{BERT} & 91.95  \% & 110M \\
             & VGG-16 \cite{VGG16}  & 90.6  \% & 130M\\
             & Eff-GNN + Word2Vec \cite{word2vec} & 91.0 \% & 124k + 610k\\
             & Eff-GNN + Word2Vec \cite{word2vec} + Image Embedding & 91.0 \% & 126k + 610k\\
                \addlinespace[1ex]
            \hline
            \hline
            \addlinespace[2ex]

     \textbf{Tobacco-3482} & VGG-16 \cite{VGG16} & 81.5\% & 130M \\
             & DocBERT \cite{DOCBERT} & 82.3 \% & 110M  \\
             & BERT \cite{BERT} & 79.0 \% & 110M \\
             & Eff-GNN+ Word2Vec \cite{word2vec} & 73.5  \% & 124k + 610k\\
             &Eff-GNN + Word2Vec \cite{word2vec} + Image Embedding & 77.5  \%& 126k + 610k\\
\bottomrule
\end{tabular}
\caption{\textbf{Classification accuracy on the Insurance, Tobacco-3482 dataset.}}
\label{effgnn_results_accuracy}
\end{table*}

\begin{table*}[ht]
\begin{tabular}{cccccc}
\toprule
    \textbf{Insurance} & \textbf{Batch Size} & \textbf{Epochs} & \textbf{Training Time} & \textbf{GPU Memory (Training)} & \textbf{Inference Time}  \\
\midrule
VGG &  32 & 28& 2.30 hours &  \shortstack[l]{7.08GB  } & 103 seconds \\

 Eff-GNN (GPU) & 32 & 50 & 3.5 mins. &\shortstack[l]{\newline \\ \newline \\470MB + 3.5 GB } & \shortstack[l]{ 0.79 seconds }  \\
  Eff-GNN (CPU) & 32 & 50 & \shortstack[l]{4.1 mins.}  &  NA & \shortstack[l]{0.79 seconds}  \\
  BERT & 16 & 15 & 6.2 hours & 10.5GB  & 40 seconds  \\
 DocBERT & 16 & 15 & 6.3 hours & 8.1GB  & \shortstack[l]{ \newline \\"40 times faster \newline\\    than  BERT"\\\cite{DOCBERT}}  \\
\bottomrule
\end{tabular}
\caption{\textbf{Memory, hardware and time required by different models on the Insurance dataset. The numbers are reported for training 4544 images and inference for 1280 Images.}}
\label{effgnn_results_resources_insurance}
\end{table*}

%% file: main/deployment.tex
\section{Applications and Deployment}
Document image classification can find many enterprise level applications in insurance companies to reduce manual review and stream line claims processes. For example, casualty injuries claims usually have multiple correspondences with hospitals and clinics, each involving documents to be processed by the insurance company. Personal automobile claims processing can use automatic document classification of letters of guarantee, purchase receipts, and others in order to auto-approve a certain percentage of auto claim reimbursements. Each year, millions of claim related document images sent to insurance companies belonging to hundreds of categorizes related to financial institutions, medical providers, or legal organizations. A scalable document classifier can assign the correct categories as meta data to a document, which can be used to route to appropriate downstream tasks. Given the scale document image classification can be deployed in an enterprise, 
an efficient framework for model training and iterations, together with an economical model hosting solutions has clear cost advantages.  
To support all those applications and achieve potential scalability , we are in the progress of deploying this trained model as an API under Container as a Service (CaaS) on AWS using Fargate managed Elastic Container Service (ECS). The efficient model size facilitates faster deployment and ease of maintenance. A standard ECS task with v4CPUs and 30G RAM can be used and the weights of the trained model can be conveniently packaged in a dockerized image and deployed at scale. 

%

%% file: main/conclusion_and_nextsteps.tex
\section{Conclusion and Future Work}
Millions of business document images, such as medical bills, attorney letters, contracts, bank statements and personal checks are processed in insurance companies to support a wide range of business workflows and applications. A scalable and efficient automation that is more intelligent than the brittle OCR-template based method is desired. 

In this paper we proposed a novel document image classification that uses graph convolution neural network to integrate text, image, and layout information of a document. We rigorously bench marked our method against the SOTA computer vision and language models on both the insurance dataset and Tobacco dataset. We also compared computing time and hardware resources required for training those models. The results showed our method is not only competitive on classification performance but also is much smaller in size therefore requires much less time and resource. This could translate to big cost advantages of hosting and deployment in real world applications. We are also working on enabling general document classification that can handle hundreds of document classes. A few options include training larger models for domain specific transfer learning, enabling few shot learning and continual learning when dynamically adding new document classes. In addition, we would like to further explore more effective document representations including more sophisticated graph representations or jointly trained layout \cite{layoutLM}.